\pdfoutput=1

\documentclass[11pt]{article}
\usepackage{acl} 
\usepackage{nert}

\usepackage{times}
\usepackage{latexsym}
\usepackage{url}
\usepackage{multirow}
\usepackage{amssymb}
\usepackage{pifont}
\usepackage{textcomp}

\usepackage[T1]{fontenc}

\usepackage[utf8]{inputenc}

\usepackage{microtype}

\usepackage{graphicx}
\usepackage{multirow}

\title{GCDT: A Chinese RST Treebank for \\ Multigenre and  Multilingual Discourse Parsing}

\author{Siyao Peng  \quad Yang Janet Liu  \quad Amir Zeldes \\
 Department of Linguistics, Georgetown University \\
  {\tt \{sp1184, yl879, amir.zeldes\}@georgetown.edu}}

\begin{document}
\maketitle
\begin{abstract}

 A lack of large-scale human-annotated data has hampered the hierarchical discourse parsing of Chinese. In this paper, we present GCDT, the largest hierarchical discourse treebank for Mandarin Chinese in the framework of Rhetorical Structure Theory (RST).
 GCDT covers over 60K tokens across five genres of freely available text, using the same  relation inventory as contemporary RST treebanks for English.
We also report on this dataset's parsing experiments, including state-of-the-art (SOTA) scores for Chinese RST parsing and RST parsing on the English GUM dataset, using cross-lingual training in Chinese and English with multilingual embeddings.

\end{abstract}

\section{Introduction}

Hierarchical discourse parsing has shown its importance in document-level natural language understanding (NLU) tasks, such as text summarization \cite{yoshida-etal-2014-dependency, goyal-eisenstein-2016-joint, xu-etal-2020-discourse, xiao-etal-2020-really, huang-kurohashi-2021-extractive} and sentiment analysis \cite{bhatia-etal-2015-better, markle-hus_improving_2017, kraus-sentiment-RST-discouse-2019, huber-carenini-2020-sentiment}.
Among discourse frameworks, 
Rhetorical Structure Theory (RST, \citealt{mann_rhetorical_1988}) is a document-level discourse analysis formalism that assumes a single-rooted, labeled constituent tree for each document. Unlike the Penn Discourse Treebank (PDTB, \citealt{miltsakaki-etal-2004-penn}), which primarily focuses on local discourse relations and for which more data exists in Chinese, 
RST builds a document tree using nested 
relations within a sentence, across sentences, and across paragraphs.
RST is thus particularly significant at the macro-level, which is more challenging for understanding discourse organization than at the micro-level \cite{jia-etal-2018-modeling, hou-2020-rhetorical-review, zhang-etal-2020-top}.

\begin{figure}[t]    
    \centering
    \includegraphics[width=0.48\textwidth]{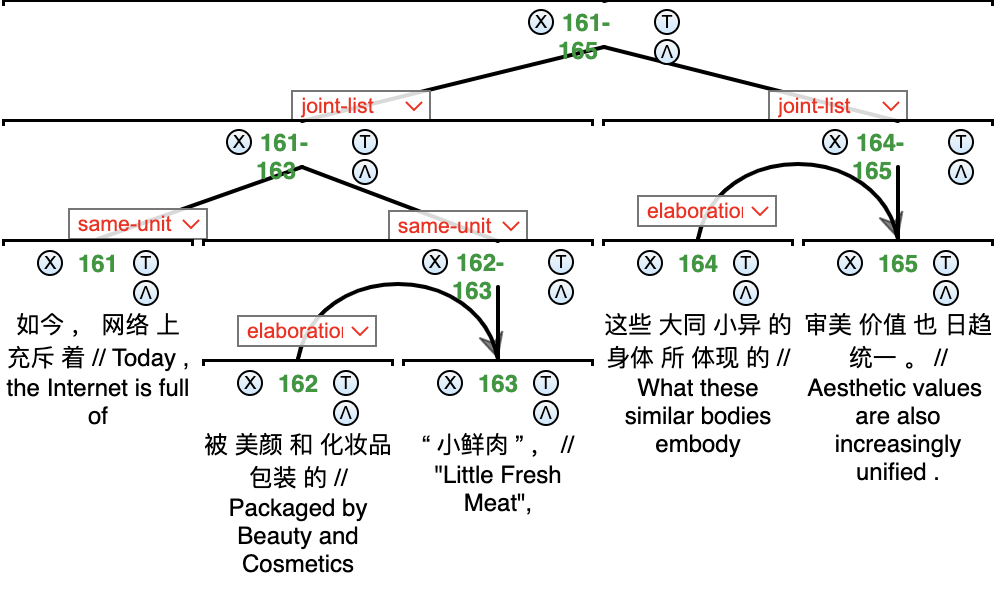}
    \caption{A RST subtree with two relative clauses annotated as \textit{elaboration-attribute} and \textit{same-unit} in \texttt{GCDT\_academic\_dingzhen} with automatic $zh\rightarrow en$ translations appended after the source Chinese texts.}
    \vspace{-5pt} 
    \label{fig:elab_attr_same_unit_dingzhen}
\end{figure}

Despite the complexity of RST and the human labor required, many new datasets have come out in the past decades \cite{zeldes-etal-2019-disrpt, zeldes-etal-2021-disrpt}, 
including English \cite{carlson-etal-2001-building, zeldes2017gum},
Basque \cite{iruskieta2013rst}, 
Bangla \cite{das-stede-2018-developing}, 
Brazilian Portuguese \cite{cardoso2011cstnews},
Dutch \cite{redeker-etal-2012-multi},
German \cite{stede-neumann-2014-potsdam},
Persian \cite{shahmohammadi2021persian},
Russian \cite{toldova-etal-2017-rhetorical},
Spanish \cite{da-cunha-etal-2011-development},
and the Spanish-Chinese parallel corpus \cite{cao-etal-2018-rst}.

However, a substantial gap remains in the availability of document-level hierarchical discourse datasets for non-European languages, particularly Chinese, of sufficient magnitude for training contemporary neural parsers. Aside from the small parallel Spanish-Chinese dataset by  \citet[see below]{cao-etal-2018-rst} with only 400+ discourse relation instances, there are no available Chinese treebanks in the RST framework. 
Thus, neither monolingual nor multilingual RST constituent parsers are trained in Chinese and cannot benefit downstream tasks. 

In this paper, we present the Georgetown Chinese Discourse Treebank (GCDT) corpus,\footnote{
The source texts, annotations, and guidelines are open-source (CC-BY) and available at
\url{https://github.com/logan-siyao-peng/GCDT}. The corpus is also searchable in the ANNIS interface \cite{zeldes-2014-annis} at \url{https://gucorpling.org/annis/\#_c=R0NEVA==}.} 
a new, freely available, multi-genre RST corpus of 50 medium to long documents for Mandarin Chinese, as the sample subtree shown in \Cref{fig:elab_attr_same_unit_dingzhen}.
The corpus covers over 60K tokens and 9K Elementary Discourse Units (EDUs).
In addition to presenting the SOTA parsing results in monolingual settings for the dataset, 
we jointly train a model with both English and Chinese datasets, testing finetuning and automatic translation-based approaches to improve performance in both Chinese and English as the target language. 
Experimenting with different monolingual and multilingual embeddings, we find that joint training and translation improve performance on the smaller Chinese and larger English datasets.
However, finetuning only helps with the smaller Chinese data. 
Finally, we show that monolingual RoBERTa embeddings outperform multilingual embeddings in applicable settings. 
Still, the best overall performance is achieved using Chinese and English data in a multilingual training regime.

\section{Previous Work}

\paragraph{RST Datasets in English and Chinese}

The English RST-DT corpus \cite{carlson-etal-2001-building} is the primary benchmark in the RST framework. The large corpus (205K tokens) includes only news articles from the Penn Treebank \citep{marcus-etal-1993-building}. Another English RST corpus is GUM \cite{zeldes2017gum}, a multi-genre corpus growing in size yearly and currently (V8.0.0) contains 180K tokens from 12 written or spoken genres.
GUM is thus slightly smaller in the token count but has a larger number of discourse relation instances due to a shorter average unit length in tokens.
Moreover, the dynamic aspect of GUM makes it different to set up benchmark scores compared to other RST corpora.
To our knowledge, this paper publishes the first set of RST parsing performances on GUM V8.0.0. 

The Spanish-Chinese parallel corpus \cite{cao-etal-2018-rst} is a small Chinese RST corpus (15K tokens) constructed for translation studies.
To support this goal, its EDUs are adjusted to align between Spanish and Chinese rather than staying faithful to the syntax of the individual languages. Its relation inventory is also distinct from inventories used for English corpora, as are the segmentation criteria used in the corpus, limiting its compatibility with other datasets.
Another older Chinese RST corpus was reported in \citet{yue2008rhetorical} with 97 news commentaries annotated. However, to our knowledge, the dataset is no longer accessible or used in RST parsing or other tasks \cite{cao-2018-dissertation-using}.

\paragraph{Other Hierarchical Chinese Discourse Datasets}

There are a few other hierarchical discourse corpora in Mandarin Chinese, but none of them annotate single-rooted RST trees for longer documents. 
The CDT-CDTB corpus \cite{li-etal-2014-building} uses connectives to build up discourse trees only within paragraphs, for 500 news documents from the Chinese Penn Treebank \cite{xue-2005-penn-chinese-treebank}. 
Not only are many of the connectives ambiguous in Chinese \cite{li-etal-2014-cross, lu2018cross}, discourse trees in CDT-CDTB are also small (only 4.5 EDUs/tree). This dataset, therefore, differs substantially from the expected structure of an RST treebank, in which EDUs are expected to be all clauses in the text with functionally motivated relation labels, such as \textit{cause} or \textit{background}.

MCDTB \cite{jiang-etal-2018-mcdtb} further utilizes a set of discourse relations to connect between paragraphs within 720 documents.
The design choice to use specific inter-paragraph-only annotations creates an interesting distinction between micro-level versus macro-level relations \cite{sporleder-lascarides-2004-combining, wang-etal-2017-two}, but also deviates from RST's fundamental idea of constructing a single tree for an entire document, in which the same inventory of labels is used for all nodes.

Moving beyond constituent-based discourse trees, 
\citet{cheng-li-2019-zero} annotated 108 scientific abstracts in their Sci-CDTB corpus using Discourse Dependency Structure  \citep[DDS;][]{hirao-etal-2013-single, morey-etal-2018-dependency}.
\citet{li-2021-unifying-discourse-dependency} further converted other Chinese discourse corpora into the DDS representation.
Even though DDS simplifies parsing
and is more similar to other linguistic annotation schemes, such as Universal Dependencies \cite{nivre-etal-2016-universal} for syntax, 
the dependency-style discourse annotation loses significant information on the ordering or scope of satellite attachments. 
For example, whether a unit with \textit{cause} and \textit{attribution} satellites means that both the cause and the result are attributed to someone, as in \Cref{sec:appendix-figure}, or that something caused an attributed statement.
In other words, when multiple discourse units modify the same nucleus, the relative importance of the satellites and their scopes are ignored.

\paragraph{Multilingual RST Parsers}

RST parsing is a task that merges a sequence of gold or predicted EDUs and forms a labeled tree structure for the entire document. 
Since RST datasets share the same unlabeled constituent tree structure, based on the principle that more prominent units should serve as nuclei to less prominent satellite units,
multilingual joint training has achieved SOTA results in multilingual RST parsing in several languages.
Translating EDUs across languages \cite{cheng-li-2019-zero, liu-etal-2020-multilingual-neural, liu-etal-2021-dmrst} and mapping word embeddings into the same space \cite{braud-etal-2017-cross, iruskieta-braud-2019-eusdisparser, liu-etal-2020-multilingual-neural, liu-etal-2021-dmrst} are two common approaches to encoding EDUs across languages in joint training. 
Among this line of work, \citet{liu-etal-2021-dmrst} presented a SOTA multilingual RST parser with a pointer-network decoder for top-down depth-first span splitting. The model uses the multilingual \textit{xlm-roberta-base} \citep{conneau-etal-2020-unsupervised} and trains jointly with six languages: English, Portuguese, Spanish, German, Dutch, and Basque. 
The current work uses the parser from \citet{liu-etal-2021-dmrst} for training between the Chinese GCDT corpus and the English GUM corpus.

\section{GCDT: Georgetown Chinese Discourse Treebank}

GCDT is an open-source multi-genre RST dataset in Mandarin Chinese. 
Following the design of GUM \cite{zeldes2017gum}, GCDT contains 50 documents, 10 from each of 5 genres which also appear in GUM:
academic articles, biographies (\textit{bio}), interview conversations, news, and how-to guides (\textit{whow}), as shown in  \Cref{tab:corpus_statistics}. 
Unlike existing Chinese discourse corpora,  GCDT focuses on building larger discourse trees for medium-to-long documents.
We select documents with an average of 1K+ tokens to provide more training data for learning higher-level discourse structures.

\begin{table}[t!bh]
\centering
\resizebox{\columnwidth}{!}{%
\begin{tabular}{c|r|r|r|l}
\hline
    \textbf{Genre}  & \textbf{\#Docs}  & \textbf{\#Toks} & \textbf{\# EDUs} & \textbf{Source}  \\ \hline 
    academic  & 10 & 14,168 & 2,033 & \url{hanspub.org/}  \\
    bio & 10 & 13,485 &2,018 & \url{zh.wikipedia.org/}  \\
    interview & 10 & 11,464 & 1,810 & \url{zh.wikinews.org/} \\
    news & 10 & 11,249 & 1,652 & \url{zh.wikinews.org/} \\
    whow & 10 & 12,539 & 2,197 & \url{zh.wikihow.com/} \\
    \hline
    Total & 50 & 62,905 & 9,710  & \\ \hline
\end{tabular}%
}
\caption{GCDT Corpus Statistics. }
\vspace{-5pt}
\label{tab:corpus_statistics}
\end{table}

\paragraph{EDU Segmentation}
Elementary Discourse Unit (EDU) segmentation is fundamental to RST. 
We deviate from previous corpora that predominately use potentially ambiguous punctuation \cite{li-etal-2014-cross} to segment EDUs, regardless of the surrounding structures. Instead, our Chinese EDU segmentation mirrors the syntactic criteria established in the English RST-DT and GUM corpora \cite{carlson2001discourse, carlson-etal-2001-building, zeldes2017gum}, largely equating EDUs with the propositional structure of clauses.
We use the Penn Chinese Treebank \cite{xue-2005-penn-chinese-treebank} as our syntactic guidelines. We first manually tokenize
according to \citet{xia2000segmentation} and conduct EDU segmentation based on parts-of-speech defined in \citet{xia2000part}.

Most notably, we segment relative clauses in GCDT, following the practice in English and other corpora \cite{carlson-etal-2001-building, zeldes2017gum, das-stede-2018-developing, cardoso2011cstnews, redeker-etal-2012-multi, toldova-etal-2017-rhetorical}.
Chinese relative clauses present a unique feature in the existing RST treebanks. To our knowledge, GCDT is the first RST corpus in any language in which prenominal relative clauses are annotated for discourse relations. 
Cross-referencing \citet{wals-90, wals-96} with languages of existing RST corpora suggests that only Basque also exhibits the Relative-Noun order found in Chinese. Yet, 
relative clauses are not segmented in the Basque RST dataset \cite{iruskieta_qualitative_2015}.
Moreover, since relative clauses intervene between Verb-Object in Chinese,
the pseudo-relation \textit{same-unit} is used to express discontinuous EDUs,
as shown in \Cref{fig:elab_attr_same_unit_dingzhen}.
Segmenting and annotating discourse relations for relative clauses is one of the reasons that GCDT has relatively short EDUs, on average 6.5 tokens\slash EDU.

\paragraph{Relation Annotation}
GCDT builds up constituent discourse trees based on gold EDUs using rstWeb \cite{zeldes-2016-rstweb}.
We use the enhanced two-level relation labels from GUM V8.0.0 with 15 coarse and 32 fine-grained relations (see \Cref{sec:appendix-label-distribution}
for relation distributions in GCDT and GUM).

\paragraph{Data Split}
We provide an 8-1-1 \texttt{train}-\texttt{dev}-\texttt{test} split per genre to facilitate future RST parsing experiments. Both human inter-annotator agreements and parsing results are assessed on the five \texttt{test} documents, with one from each of the five genres. 

\paragraph{Inter-Annotator Agreement (IAA)}
We evaluate agreement on the five \texttt{test} documents to obtain human ceiling scores for parser performance. 
One Chinese native-speaker linguist annotated the entire corpus, and another read the guidelines and conducted independent EDU segmentation. 
We measured segmentation agreement, adjudicated segmentation between the two annotators, and then separately annotated relation trees on gold EDUs to measure relation agreement.
We also release the double annotations in GCDT for future experiments on annotation disagreements.
On segmentation, we obtained a token-wise agreement of 97.4\% and
Cohen's $\kappa=$0.89. 
The agreements on 
micro-averaged original Parseval F1 of Span, Nuclearity, and Relation are 84.27, 66.15, and 57.77 respectively. 
The IAA of GCDT is similar to that of the English RST-DT benchmark -- 78.7, 66.8, and 57.1 -- when evaluated using the original Parseval \cite{morey-etal-2017-much}. 
The results show that the GCDT annotation agreement is highly satisfactory even though the documents are much longer and exhibit more genre diversity than RST-DT.

\begin{table*}[t!hb]
\centering
\resizebox{\textwidth}{!}{%
\begin{tabular}{c|cccc|cccc}
\hline
corpus & monolingual embedding 
& Span & Nuc & Rel 
& multilingual embedding 
& Span & Nuc & Rel  \\
\hline
\multirow{3}{*}{\begin{tabular}[x]{@{}c@{}}GCDT\end{tabular}}
& \textit{bert-base-chinese} 
& 73.15\textpm0.53 & 55.71\textpm0.66 & 50.81\textpm0.65
& \textit{bert-base-multilingual-cased} 
& 67.34\textpm1.32 & 47.66\textpm0.73 & 43.97\textpm0.93 \\
& \textbf{\textit{hfl/chinese-roberta-wwm-ext}}
& \textbf{75.51\textpm0.68} & \textbf{57.08\textpm0.81} & \textbf{51.76\textpm0.97}  
& \textbf{\textit{xlm-roberta-base}} 
& \textbf{74.35\textpm0.54} & \textbf{54.17\textpm1.20} & \textbf{50.45\textpm1.09} \\
\hline
\multirow{3}{*}{\begin{tabular}[x]{@{}c@{}}GUM-5\end{tabular}}
& \textit{bert-base-cased} 
& 64.61\textpm1.42 & 49.58\textpm1.51 & 40.43\textpm1.56
& \textit{bert-base-multilingual-cased} 
& 64.52\textpm2.68 & 51.63\textpm2.07 & 44.96\textpm1.46 \\
& \textbf{\textit{roberta-base}} 
& \textbf{73.85\textpm0.70} & \textbf{58.95\textpm0.79} & \textbf{50.35\textpm1.18} 
& \textbf{\textit{xlm-roberta-base}} 
& \textbf{72.45\textpm0.97} & \textbf{56.78\textpm0.80} & \textbf{47.69\textpm0.88} \\
\hline
\multirow{3}{*}{\begin{tabular}[x]{@{}c@{}}GUM-12\end{tabular}}
& \textit{bert-base-cased} 
& 60.93\textpm0.63 & 47.92\textpm0.62 & 40.20\textpm0.40  
& \textit{bert-base-multilingual-cased} 
& 64.47\textpm0.50 & 50.69\textpm0.32 & 43.25\textpm0.35 \\
& \textbf{\textit{roberta-base}} 
& \textbf{68.59\textpm0.58} & \textbf{55.32\textpm0.27} & \textbf{46.29\textpm0.46}  
&  \textbf{\textit{xlm-roberta-base}}
& \textbf{66.12\textpm0.59} & \textbf{52.58\textpm0.52} & \textbf{45.06\textpm0.45}  \\
\hline
\end{tabular}%
}
\caption{Monolingual parsing results on the test sets of GCDT, GUM-5, and GUM-12 with Chinese, English, and multilingual BERT and RoBERTa embeddings (mean\textpm std over five runs). }
\label{tab:result_monolingual}
\end{table*}

\begin{table*}[t!hb]
\centering
\resizebox{\textwidth}{!}{%
\begin{tabular}{lccc|lccc}
\hline
Experiment & Span & Nuc & Rel 
& Experiment & Span & Nuc & Rel 
\\
\hline
\multicolumn{4}{c|}{\textbf{Train on GCDT+GUM-5 and Dev/Test on GCDT}} 
& \multicolumn{4}{c}{\textbf{Train on GUM-5+GCDT and Dev/Test on GUM-5}}
\\
joint training w/ XLM RoBERTa
& 74.24\textpm0.48 & 56.68\textpm0.86 & 52.21\textpm0.83
& joint training w/ XLM RoBERTa
& 72.56\textpm0.71 & 60.63\textpm0.43 & 52.57\textpm0.77
\\
+finetuning w/ XLM RoBERTa
& 76.97\textpm0.32 & 57.94\textpm0.82 & 53.38\textpm0.51
& +finetuning w/ XLM RoBERTa
& 73.44\textpm0.36 & 59.40\textpm0.56 & 50.57\textpm0.97
\\
+en$\rightarrow$zh trans. w/ 
XLM RoBERTa
& 74.80\textpm0.78 & 56.58\textpm0.98 & 51.18\textpm1.15
& +zh$\rightarrow$en trans. w/ 
XLM RoBERTa
& 72.21\textpm1.11 & 60.07\textpm1.25 & 52.32\textpm1.05
\\
\textbf{+en$\rightarrow$zh trans. w/ 
ZH RoBERTa}
& \textbf{77.66\textpm0.42} & \textbf{59.29\textpm0.59} & \textbf{54.66\textpm0.76}
& \textbf{+zh$\rightarrow$en trans. w/ 
EN RoBERTa}
& \textbf{74.73\textpm0.40} & \textbf{62.65\textpm0.72} & \textbf{54.32\textpm0.82} 
\\
\hline
\multicolumn{4}{c|}{\textbf{Train on GCDT+GUM-12 and Dev/Test on GCDT}}  
& \multicolumn{4}{c}{\textbf{Train on GUM-12+GCDT and Dev/Test on GUM-12}} 
\\
joint training w/ 
XLM RoBERTa
& 74.33\textpm0.49 & 57.24\textpm0.99 & 52.61\textpm1.13
& joint training w/ 
XLM RoBERTa
& 70.32\textpm0.37 & 57.49\textpm0.73 & 49.14\textpm0.34
\\
\textbf{+finetuning w/ 
XLM RoBERTa}
& \textbf{76.95\textpm0.65} & \textbf{59.40\textpm0.64} & \textbf{55.28\textpm0.23}
& +finetuning w/ 
XLM RoBERTa
&  66.00\textpm0.24  &  53.13\textpm0.22  &  45.47\textpm0.42
\\
+en$\rightarrow$zh trans. w/ 
XLM RoBERTa
& 73.99\textpm0.79 & 56.31\textpm1.43 & 51.51\textpm1.34
& +zh$\rightarrow$en trans. w/ 
XLM RoBERTa
& 70.28\textpm0.55 & 57.63\textpm0.55 & 49.26\textpm0.39
\\
+en$\rightarrow$zh trans. w/ 
ZH RoBERTa
& 78.11\textpm0.39 & 59.42\textpm0.90 & 54.41\textpm1.23
& \textbf{+zh$\rightarrow$en trans. w/ 
EN RoBERTa}
& \textbf{71.41\textpm0.47} & \textbf{59.17\textpm0.35} & \textbf{50.63\textpm0.48}
\\
\hline
  \end{tabular}%
} 
\caption{Multilingual parsing results with finetuning and automatic translation  on the test sets of GCDT+GUM combinations with highest-performing Chinese (ZH), English (EN), and multilingual (XLM) RoBERTa embeddings. 
}
\label{tab:result_multilingual}
\end{table*}

\begin{table*}[t!hb]
\centering
\resizebox{0.9\textwidth}{!}{%
\begin{tabular}{c|ccc|ccc|ccc|ccc}\hline
\multirow{2}{*}{\begin{tabular}[x]{@{}c@{}}Genre\end{tabular}}
&  \multicolumn{3}{c|}{Trained on GCDT}
&  \multicolumn{3}{c|}{\begin{tabular}[x]{@{}c@{}}Trained on GCDT+GUM-5 \\ w/ zh$\rightarrow$en trans.\end{tabular}}
&  \multicolumn{3}{c|}{\begin{tabular}[x]{@{}c@{}}Trained on GCDT+GUM-12 \\ w/ zh$\rightarrow$en trans.\end{tabular}}
& \multicolumn{3}{c}{\begin{tabular}[x]{@{}c@{}}Human \\ Agreement \end{tabular}}
\\
\hline
& Span & Nuc & Rel 
& Span & Nuc & Rel 
& Span & Nuc & Rel 
& Span & Nuc & Rel 
\\\hline
academic  
&  74.64	& 54.07	& 48.33	
& 72.25 & 47.37 & 43.54 
& 75.12 & 51.20 & 44.98  
& 80.38 & 59.33 & 49.76  
\\
bio  
& 72.87	& 54.26	& 52.71	
& 74.81 & 57.75 & 53.49 
& 77.52 & 59.69 & 55.43  
& 81.57 & 63.92 & 55.69 

\\
interview  
& 74.68	& 56.33	& 52.53	
& 80.38 & 61.39 & 55.70 
& 77.85 & 56.96 & 48.73  
& 83.55 & 62.50 & 54.61 
\\
news  
&  76.63	& 56.52	& 50.54
& 83.15 & 64.13 & 57.07 
& 78.80 & 60.33 & 54.35  
& 80.98 & 61.96 & 54.35 
\\
whow  
&  77.89	& 57.76	& 54.79   
& 80.20 & 66.34 & 62.71 
& 80.20 & 65.68 & 61.06  
& 91.99 & 77.70 & 69.34 
\\
\hline
Overall  
& 75.45 & 55.85 & 52.07 
& 77.97 & 59.71 & 55.04 
& 78.06 & 59.44 & 53.87  
& 84.27 & 66.15 & 57.77 
\\\hline 
\end{tabular}
}
\caption{GCDT genre-wise performances on sample models trained on GCDT, as well as translation-augmented GCDT+GUM-5 and GCDT+GUM-12 combinations using  \textit{hfl/chinese-roberta-wwm-ext}.}
\label{tab:genre-wise-gcdt-performance}
\vspace{-8pt} 
\end{table*}

\section{Experiments}
\label{sec:experiments}

We present benchmark results on GCDT using the SOTA multilingual parser, DMRST \cite{liu-etal-2021-dmrst}.
Results are shown in two experimental settings: \textit{\textbf{monolingual}} training using only one dataset (either Chinese GCDT or English GUM V8.0.0) and \textit{\textbf{multilingual}} training using data from both corpora (GCDT+GUM).
Besides directly combining corpora from the two languages, we also experiment with finetuning and automatic EDU-wise translation. 
We use the same set of hyperparameters as reported in \citet{liu-etal-2021-dmrst}. 
Similarly, we also report monolingual and multilingual parsing performance on GUM V8.0.0.

\paragraph{Datasets}
Cross-genre adaptability remains a bottleneck in RST parsing \cite{nishida-2022-out-of-domain-discourse-parsing, atwell-etal-2021-discourse}.
To isolate cross-lingual versus cross-genre influences, 
we conduct monolingual and multilingual experiments using the following data compositions: 
1) \textbf{GCDT}: 50 Chinese documents from 5 genres; 
2) \textbf{GUM-12}: 193 English documents from 12 genres; 
3) \textbf{GUM-5}: 99 GUM documents from the same 5 genres in GCDT.

\paragraph{Language Models}
We test different monolingual and multilingual BERT and RoBERTa embeddings (see \Cref{sec:appendix-lm} for details).

\paragraph{Metrics}
We use the 15 coarse relation classes shared between GCDT and GUM and follow the recommendation of \citet{morey-etal-2017-much} to use the original Parseval micro-averaged F1 for Span, Nuclearity (Nuc), and Relation (Rel).

\paragraph{Multilingual Training Setups}
In addition to training with combinations of the original GCDT and GUM datasets using multilingual embeddings (see \Cref{sec:appendix-multilingual-data-split} for specific data partitions used in the GCDT+GUM-combined experiments), 
we also experiment with two techniques to improve performances on both target datasets.
Specifically, to improve on Chinese GCDT:

1) \textbf{Finetuning}: 
we first train models with both English and Chinese data and then continue training only on the training partition of the target dataset (i.e.,~GCDT).

2) \textbf{Automatic EDU-wise Translation}: 
we use GoogleTranslator\footnote{
\url{https://github.com/nidhaloff/deep-translator}} to automatically translate EDUs from the other dataset to the target language (i.e.,~EDU-wise en$\rightarrow$zh translations of GUM) and train on the original GCDT and translated GUM data. The advantage of the translation approach is that we can replace the multilingual embeddings with higher-performing monolingual embeddings.

\section{Results}
\label{sec:results}

We present monolingual and multilingual results on GCDT and GUM in 
\Cref{tab:result_monolingual,tab:result_multilingual}, as well as genre-wise performance on GCDT in \Cref{tab:genre-wise-gcdt-performance}.

\paragraph{Monolingual Parsing}
Similar to previous observations
\cite{staliunaite-iacobacci-2020-compositional,
naseer-2021-ieee-empirical-comparison-bert-roberta,
tarunesh2021trusting}, 
\Cref{tab:result_monolingual} shows that RoBERTa outperforms BERT in both languages.
Monolingual RoBERTa embeddings achieve the best performance when training with monolingual data,
e.g.,~\textit{hfl/chinese-roberta-wwm-ext} obtained 51.76 on the relation level on GCDT.

\paragraph{Multilingual Parsing}
Our multilingual parsing experiments include joint training, finetuning, and automatic EDU-wise translation. 
Based on the monolingual results, we use the best-performing multilingual embedding \textit{xlm-roberta-base} \cite{conneau-etal-2020-unsupervised} 
with the GCDT+GUM-combined multilingual data.
Different aspects of the multilingual parsing results are shown in 
\Cref{tab:result_multilingual}.

Firstly, joint training outperformed monolingual results in all three test scenarios: GCDT, GUM-5, and GUM-12. For example, training on GCDT+GUM-12 using XLM RoBERTa achieved an F\_Rel of 52.61 on GCDT, higher than the 50.45 trained with only GCDT, and the same embedding.
Secondly, more genres from GUM (GCDT+GUM-12) achieved better performance than training only using the same genres (GCDT+GUM-5) when tested on GCDT.
Thirdly, pretraining on the GCDT+GUM-combined training sets and training on the training set of the target corpus improves performance on Chinese GCDT but deteriorates on the English GUM.
We hypothesize that with more English training data available, there is less headroom for improvement. In contrast, finetuning for the smaller Chinese dataset added to the comparatively little information available to the parser.
Lastly, results show that augmenting with automatic translation and using monolingual embeddings achieved the best performance on three of the four test scenarios, while the best result on GCDT was achieved by training together with GUM-12 and finetuning on GCDT.

\paragraph{Genre-wise Analysis} 
We further select three models trained in the monolingual GCDT and translation-augmented scenarios, GCDT+GUM-5 and GCDT+GUM-12, using the Chinese RoBERTa embedding \cite{hfl-iflytek-chinese-roberta-bert}.
\Cref{tab:genre-wise-gcdt-performance} provides per-genre parsing results of the models on the five \texttt{test} genres.
On the one hand, the average performance on how-to guides (\textit{whow}) is much higher than \textit{academic} articles for both models and humans. This demonstrates a good human-model alignment regarding which genre is the hardest or easiest (cf.~\citealt{ZeldesSimonson2016}).
On the other, model results are the farthest from the human ceiling scores on the highest performing \textit{whow} genre.  
We hypothesize that characteristics of genres triggered the different performances.
Future multi-genre experiments could be conducted across datasets to study out-of-domain effects in multilingual RST parsing scenarios.

\section{Conclusion}
This paper presents GCDT, the largest RST dataset for Mandarin Chinese, which closely follows established RST guidelines and is highly comparable to existing English RST corpora.
Besides evaluating annotation quality and establishing SOTA results on this dataset in monolingual settings,
we also jointly train on a similar English corpus---GUM---and demonstrate that multilingual training and automatic EDU translation boost parser performance.
However, finetuning is only helpful when targeting the smaller Chinese dataset.
We further conduct per-genre analyses and show that parsing performance varies widely between some genres but less between others.
We hope that this dataset can alleviate the lack of training resources for hierarchical discourse parsing in Chinese and facilitate multilingual and multi-genre RST parsing, as well as other downstream NLP tasks.

\section*{Acknowledgements}
We thank Nianwen Xue for providing insights on Chinese syntax, which helped refine the EDU segmentation guidelines.
We also thank the anonymous reviewers and Nathan Schneider for their insightful comments.

\bibliography{main}
\bibliographystyle{acl_natbib}

\clearpage
\appendix
\section{Label Distributions}
\label{sec:appendix-label-distribution}

\Cref{tab:relation_distribution} gives descriptive statistics of the distribution of relations in GCDT and numbers for comparison from the GUM corpus, which uses the same inventory of relations and covers all and more genres in the GCDT dataset.

\begin{table}[ht]
\centering
\resizebox{0.48\textwidth}{!}{%
\begin{tabular}{crr}

Relation Name & GCDT\% & GUM\% \\ 
\hline
\multicolumn{3}{c}{Nucleus-Satellite Relations} \\
\hline
elaboration-attribute & 7.71\% & 4.60\% \\
attribution-positive & 4.37\% & 3.08\% \\
elaboration-additional & 4.25\% & 4.81\% \\
explanation-evidence & 4.15\% & 2.08\% \\
context-background & 2.89\% & 2.66\% \\
context-circumstance & 2.69\% & 2.40\% \\
organization-preparation & 2.36\% & 1.83\% \\
causal-cause & 1.98\% & 1.63\% \\
organization-heading & 1.78\% & 1.49\% \\
contingency-condition & 1.77\% & 1.67\% \\
adversative-concession & 1.68\% & 2.04\% \\
purpose-goal & 1.54\% & 1.63\%  \\
restatement-partial & 1.32\% & 1.13\% \\ 
evaluation-comment & 1.15\% & 2.29\% \\
mode-means & 1.09\% & 0.55\% \\
explanation-justify & 0.87\% & 1.60\% \\
causal-result & 0.87\% & 1.54\% \\
adversative-antithesis & 0.58\% & 1.47\% \\
mode-manner & 0.52\% & 0.89\% \\
topic-question & 0.44\% & 1.10\% \\
organization-phatic & 0.27\% & 1.37\% \\
attribution-negative & 0.23\% & 0.30\% \\
purpose-attribute & 0.21\% & 0.87\% \\
explanation-motivation & 0.2\% & 0.71\% \\
topic-solutionhood & 0.01\% & 0.20\% \\
\hline
\multicolumn{3}{c}{Multi-Nucleus Relations} \\
\hline
joint-list & 22.28\% & 12.90\% \\
same-unit & 18.69\% & 4.71\% \\
joint-sequence & 4.99\% & 4.41\% \\
joint-other & 4.83\% & 4.48\% \\
adversative-contrast & 3.32\% & 2.40\% \\
joint-disjunction & 0.64\% & 1.13\%  \\
restatement-repetition & 0.32\% & 1.82\% \\
\hline
\end{tabular}%
}
\caption{Distribution of 32 relations (15 classes, including \textit{same-unit}) in GCDT and GUM V8.0.0. }
\label{tab:relation_distribution}
\end{table}

\section{Specific PLMs Used in the Experiments}
\label{sec:appendix-lm}

\Cref{tab:lang-embeddings} shows the Chinese, English, and multilingual BERT and RoBERTa pretrained language models used in the experiments described in \S\ref{sec:experiments}. 

\begin{table}[ht]
\centering
\resizebox{0.48\textwidth}{!}{%
\begin{tabular}{c|l}
\hline
 Type & Details \\ 
 \hline
\multirow{3}{*}{\begin{tabular}[x]{@{}c@{}}\textbf{BERT}\end{tabular}}
& Chinese: \textit{bert-base-chinese} \cite{devlin-etal-2019-bert} \\
& English: \textit{bert-base-cased} \cite{devlin-etal-2019-bert}  \\
& Multilingual: \textit{bert-base-multilingual-cased} \cite{devlin-etal-2019-bert} \\ \hline

\multirow{3}{*}{\begin{tabular}[x]{@{}c@{}}\textbf{RoBERTa}\end{tabular}}
& Chinese: \textit{hfl/chinese-roberta-wwm-ext} \cite{hfl-iflytek-chinese-roberta-bert} \\
& English: \textit{roberta-base} \cite{liu2019roberta}    \\  
& Multilingual: \textit{xlm-roberta-base} \cite{conneau-etal-2020-unsupervised} \\ 
\hline
\end{tabular}%
}
\caption{An overview of pretrained BERT and RoBERTa language models used in the experiments. }
\label{tab:lang-embeddings}
\end{table}

\section{A Fragment of RST Annotation in GCDT}
\label{sec:appendix-figure}

\begin{figure}[ht]    
    \centering
    \includegraphics[width=0.40\textwidth]{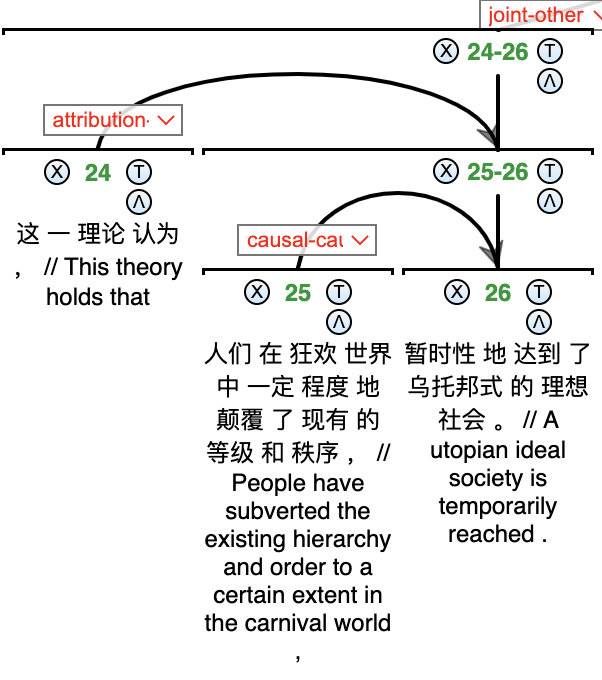}
    \caption{A RST subtree with  \textit{attribution-positive} scoping over \textit{causal-cause} from \texttt{GCDT\_academic\_dingzhen} with automatic $zh\rightarrow en$ translation.}
    \label{fig:cause_attribution_dingzhen}
\end{figure}

\Cref{fig:cause_attribution_dingzhen} presents a relation hierarchy of \textit{attribution-positive} scoping over \textit{causal-cause}. 


\section{Data Splits for Multilingual Experiments}
\label{sec:appendix-multilingual-data-split}

\Cref{tab:multilingual-combined-data-partition} presents the \texttt{train/dev/test} splits when jointly training with GCDT and GUM in multilingual experiments.

\begin{table}[ht]
\centering
\resizebox{0.5\textwidth}{!}{%
\begin{tabular}{c|ccc} \hline
& \begin{tabular}[x]{@{}l@{}}\texttt{train}: GCDT+GUM \\ \texttt{dev/test}: GUM\end{tabular} 
& \begin{tabular}[x]{@{}l@{}}\texttt{train}: GCDT+GUM \\ \texttt{dev/test}: GCDT\end{tabular} 
\\
\hline
\texttt{train} 
&   \begin{tabular}[x]{@{}c@{}} GUM-train \\ + GCDT-train \\ + GCDT-dev \end{tabular} 
& \begin{tabular}[x]{@{}c@{}}GCDT-train \\ + GUM-train \\ + GUM-dev \end{tabular} 
\\
\hline
\texttt{dev}   & GUM-dev & GCDT-dev \\
\hline
\texttt{test} & GUM-test & GCDT-test \\ \hline
\end{tabular}
}
\caption{An overview of the \texttt{train/dev/test} splits of GCDT and GUM used for training in the multilingual experiments. }
\label{tab:multilingual-combined-data-partition}
\end{table}

\end{document}